\documentclass[11pt,a4paper]{article}
\usepackage[hyperref]{conll2018}
\usepackage{times}
\usepackage{latexsym}
\usepackage{graphicx}
\usepackage{url}
\graphicspath{ {img/} }
\usepackage{amsmath,amssymb}
\usepackage{algorithm2e}
\usepackage{amsmath}
\usepackage{subcaption}
\DeclareMathOperator*{\argmax}{arg\,max}

\aclfinalcopy 

\title{Comparing Models of Associative Meaning: An Empirical Investigation of Reference in Simple Language Games}

\author{Judy Hanwen Shen   \qquad Matthias Hofer  \qquad  Bjarke Felbo  \qquad  Roger Levy \\
Massachusetts Institute of Technology \\
77 Massachusetts Avenue\\
Cambridge, MA 02139\\
  {\tt \{judyshen, mhofer, felbo, rplevy\}@mit.edu}}

\date{}

\begin{document}
\maketitle
\begin{abstract}
Simple reference games \citep{wittgenstein50philosophical} are of central theoretical and empirical importance in the study of situated language use.  Although language provides rich, compositional truth-conditional semantics to facilitate reference, speakers and listeners may sometimes lack the overall lexical and cognitive resources to guarantee successful reference through these means alone.  However, language also has rich associational structures that can serve as a further resource for achieving successful reference. Here we investigate this use of associational information in a setting where \emph{only} associational information is available: a simplified version of the popular game Codenames. Using optimal experiment design techniques, we compare a range of models varying in the type of associative information deployed and in level of pragmatic sophistication against human behavior.  In this setting we find that listeners' behavior reflects direct bigram collocational associations more strongly than word-embedding or semantic knowledge graph-based associations and that there is little evidence for pragmatically sophisticated behavior by either speakers or listeners of the type that might be predicted by recursive-reasoning models such as the Rational Speech Acts theory. 
These results shed light on the nature of the lexical resources that speakers and listeners can bring to bear in achieving reference through associative meaning alone.
\end{abstract}

\section{Introduction}
\label{sec:related_work}

In his 1953 book \textit{Philosophical Investigations}, Wittgenstein makes the argument for studying simple reference games to learn about the nature of language \citep{wittgenstein50philosophical}. Various applications of this idea in different fields, including linguistics \citep{pietarinen2007game}, cognitive science \citep{frank2012predicting}, artificial intelligence \citep{lazaridou2016multi}, and behavior-based robotics \citep{steels1997synthetic} have validated this fundamental insight and demonstrated the theoretical and empirical importance of studying language learning and use in simplified contexts. Here we describe a novel framework that uses a simple reference game to study the \textit{semantic resources} speakers and listeners use to facilitate reference. In particular, placing strong constraints on word choice and modes of interaction allows us to better isolate specific aspects that contribute towards the complexity of natural language semantics. Language provides a multitude of different resources for its users to cooperatively achieve reference. In particular, language provides truth-conditional semantic structures. These information structures are characterized in terms of their logical truth conditions and can be precisely stated using formal logic. Across many cases, however, successful reference cannot be guaranteed through these means alone. Another possible source of semantic information are \textit{associative resources} (e.g. the meaning associations of `nurse' with `female nurse' rather than `male nurse'). The question of how to best formally characterize these rich associative structures to adequately account for our linguistic abilities is still largely unresolved. 

We compare the performance of different models in accounting for human behavior in a simple reference game, a modified version of the popular board game \textit{Codenames}. Crucially, in this setting, only associational information is available. To allow us to additionally address questions about possible pragmatic effects when playing the game, our models are formulated in the context of the Rational Speech Act (RSA) framework \citep{frank2012predicting}. The candidate models of human semantic reasoning we consider involve different types of associative resources and different degrees of pragmatic sophistication by speaker and listener. The models correspond to qualitatively different sources of information, including {collocations}, {distributional similarity} across contexts, {topic similarity}, or common-sense {conceptual relatedness}. 

In the closest predecessor to our work, \citet{xu_inference} used observational data from the television game \textit{Password}, where the goal is to guess a target word on an associated cue word freely generated, to model whether speaker and listener alignment based on their differential reliance on forward vs backward word associations (estimated using the experimental norms of \citealp{nelson-etal:2004usf-norms}).  They found that similar mixtures of forward and backward associations best explained both speaker and hearer behaviors, suggesting game participants are well calibrated and cooperative with another, but did not investigate the nature of the lexical knowledge accounting for the associations underlying participant behavior.

In this paper, we construct a simplified reference game involving word associations where constrained sets of potential reference clues words and reference target words are provided. We construct a variety of different semantic association measures and conduct a series of experiments to test which source of information humans use. Furthermore, we combine these measures with the RSA framework to derive predictions about pragmatic behavior on the task.

\section{Experimental methods}
\begin{figure*}[h!]
\centering
\includegraphics[width=1.0\textwidth]{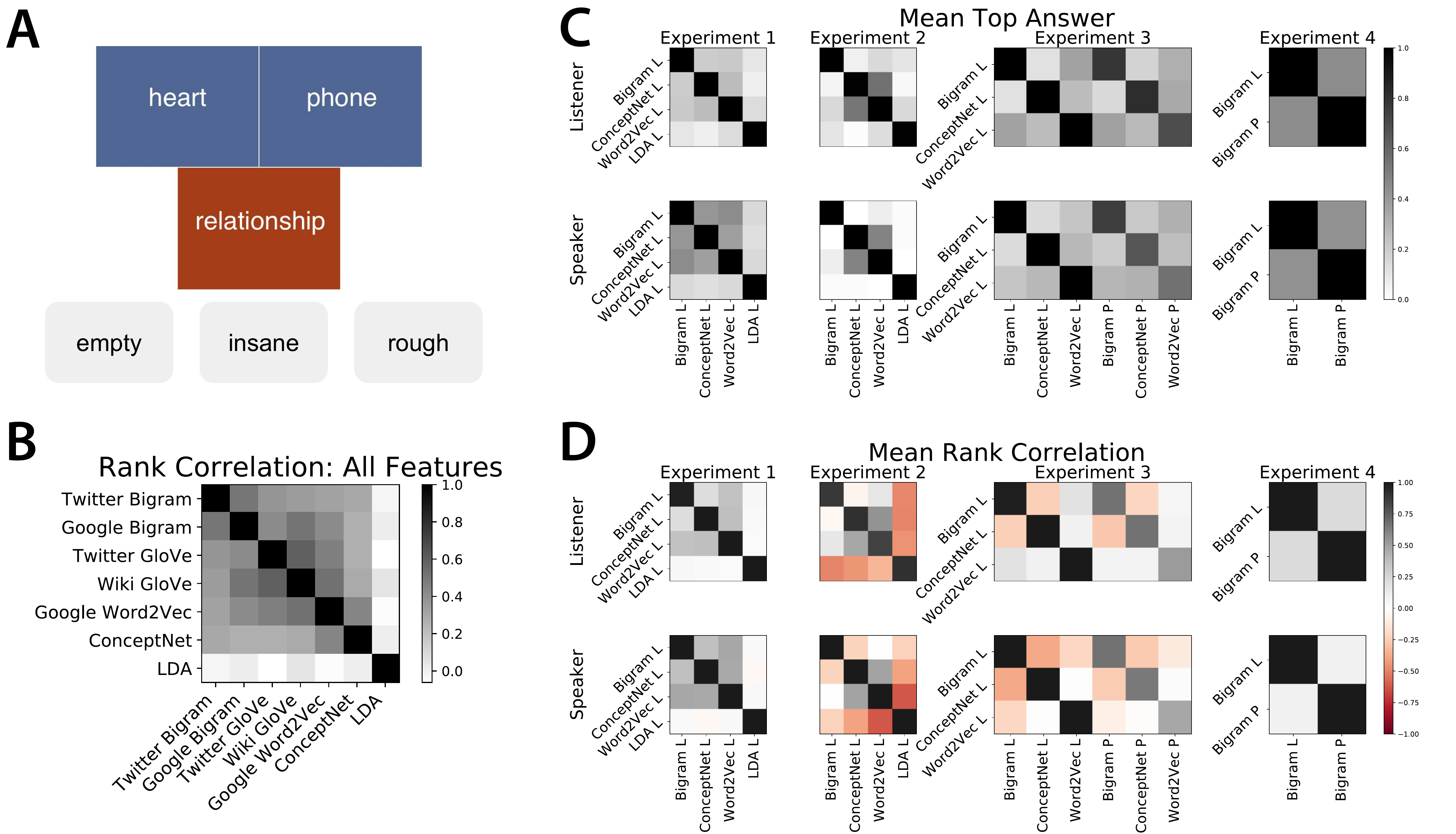}
\caption{\textbf{A.} Example of an experimental display in the speaker condition. The choice is between three adjectives (gray) to best communicate the (blue) target words while avoiding the (red) non-target words. \textbf{B.} Rank correlation between semantic association scores on the entire set of 5,000 noun--adjective pairs that all experiments draw from. \textbf{C.} Each cell shows the mean top answer matches between model pairs for the configurations used in a particular experiment (speaker and listener side). \textbf{D.} Here each cell shows the mean Spearman's correlation coefficient between model pairs for the configurations used in a particular experiment (speaker and listener side).}
\label{fig:model_macro}
\end{figure*}

We create a simplified version of the board game Codenames \citep{codenames} where the objective is for a speaker to select a clue word that allows a listener to correctly identify a set of target words. Subjects play one scenario per turn. A scenario consists of a set of \textit{codenames} drawn from a list of 50 common nouns, two of which are \textit{targets} while the remaining nouns are \textit{non-targets}. While both listeners and speakers always see the set of codenames, only the speaker knows which nouns are targets and non-targets (see Figure \ref{fig:model_macro}A, the listener views three identical black and white cards). We refer to any combination of two nouns as a \textit{noun pair}. Each scenario also contains a set of \textit{clues} drawn from 100 descriptive adjectives. Throughout the paper, we will interchangeably refer to codenames as nouns and to clues as adjectives. A \textit{configuration} is a scenario that additionally includes an \textit{index}, either indicating the target noun pair (speaker configuration) or the adjective that was provided to the listener (listener configuration). Thus, while scenarios are just lists of adjectives and nouns, there are $\binom{\# codenames}{2}$ possible speaker configurations and $\# clues$ possible listener configurations. 
 
Speakers and listeners participated in separate versions of the experiment, but were told that they would be teamed up with another player to increase engagement with the task. Subjects were either in the speaker role or in the listener role. On each trial, depending on their role, they were either given a speaker configuration or a listener configuration, that is, a scenario plus corresponding index. The speaker's task is to select a single adjective to best communicate the target noun pair without including any non-target nouns. For the listener task, participants are given an adjective and asked to select the two nouns that the adjective most likely refers to. To quantify the difficulty of a particular configuration for participants, we additionally asked them to rate how confident they were in their answer on a scale from $1$ (least confident) to $5$ (most confident). We conducted four experiments for which we recruited a total of $1460$ subjects on Amazon's Mechanical Turk platform. Each subject completed 20 different configurations, lasting approximately $7-10$ minutes and were paid a fixed amount of $\$0.60$ for their participation. We make all data and analysis code available \footnote{\url{https://github.com/heyyjudes/codenames-language-game/}}.

\subsection{Modeling word choice}
\label{sec:rsa_equations}
Following previous work on linguistic reasoning as Bayesian inference, subjects' choices for a given configuration are modeled using the Rational Speech Acts (RSA) model. In the model, a pragmatic listener $L_1$ reasons recursively about a pragmatic speaker $S_1$ that in turn reasons about a literal listener $L_0$. Referents are noun pairs, $p$, and utterances are adjectives, $a$. We assume a uniform prior over possible adjectives and over possible noun pairs. Communication costs are set to $0$. Assuming that adjectives are chosen in proportion to the degree of semantic association between a noun pair and an adjective, denoted as $s_{p,a}$, we obtain the set of simplified equations shown in Table \ref{tab:rsa_eq}. Experiments 1-3 in the following sections will use $P_{L_0}(p|a)$ and $P_{S_0}(p|a)$ while Experiment 4 compares the two aforementioned literal agents with the pragmatic versions using $P_{L_1}(p|a)$ and $P_{S_1}(p|a)$. 

\begin{table}
\small
\begin{center}
\begin{tabular}{ll}
\textbf{Listener} & \textbf{Speaker} \\[4pt]
\hline\\[-4pt] 
$P_{L_0}(p|a) \propto s_{p,a}$                       & $P_{S_0}(a|p) \propto s_{p,a}$\\[2pt]
$P_{S_1}(a|p) \propto [P_{L_0}(p|a)]^{\alpha}$ 		 & $P_{L_1}(p|a) \propto [P_{S_0}(a|p)]^{\alpha}$\\[2pt]
$P_{L_1}(p|a) \propto P_{S_1}(a|p)$                  & $P_{S_1}(a|p) \propto P_{L_1}(p|a)$ \\[2pt]
\end{tabular}
\end{center}
\caption{\label{tab:rsa_eq} RSA equations for constructing literal and pragmatic models from semantic metrics.}
\end{table}

\subsection{Modeling semantic association strength}
Our primary interest is understanding how people reason about the semantic relatedness of arbitrary noun--adjective pairings, formally expressed as different semantic association metrics $s_{p,a}$. Unlike in previous applications of RSA, where an utterance is either true of a particular referent or not, the relation between nouns and adjectives in the present setting is one of \textit{associative strength}: an adjective can fit a noun to different degrees \citep{perea2002effects}.\footnote{The adjective `dirty', for instance, is more strongly associated with the noun `pig' than with the noun `slate'. In contrast, `slate' is more strongly associated with the adjective's antonym `clean', likely owing the widespread collocation `clean slate'.} Here we consider four different types of models to quantify the semantic relatedness $s_{n,a}$ between a noun $n$ and an adjective $a$. This measure is extended to cover noun pairs $p=\{n_1, n_2\}$ by product aggregation: $s_{p,a} = s_{n_1,a} \cdot s_{n_2,a}$.

\subsubsection{Bigram semantic association}
\label{sec:bigram_association}
The first metric we consider is derived from the bigram co-occurrence counts of noun--adjective pairs $z_{n,a}$, describing how relevant an adjective $a \in A$ is for a noun $n \in N$. We create one set of these relationships using Google Ngram probabilities averaged across the years 1990 to 2000 \cite{michel2011quantitative}. A comparison set is obtained from a real-world corpus containing $30$B messages from Twitter. The semantic association is computed as: 
\begin{align}
s_{n,a} &= \frac{P(a|n)}{P(a)}
\label{eq:semantic_association_sub}
\end{align}
Eqn.~\ref{eq:semantic_association_sub} captures how often an adjective occurs with a noun while normalizing for the frequency of the adjectives. 

\subsubsection{Vector embedding cosine distance}
\label{sec:vector_dist}
Global Vectors for Word Representation (GloVe) \cite{pennington2014glove} and skip-gram model trained vectors (Word2Vec) provide vector representations for words that encompass semantic and linguistic similarity. We examine the Twitter GloVe set ($d = 200$), the Wiki-GigaWord GloVe set of ($d = 200$) \cite{pennington2014glove}, and Google News Word2Vec vectors ($d = 300$) \cite{mikolov2013distributed}. To calculate noun--adjective similarities, we compute cosine distance between each noun--adjective pair's vector embeddings.

\subsubsection{ConceptNet5 similarity}
\label{sec:concept_sim}
ConceptNet combines knowledge from a variety of sources, including Wiktionary~\footnote{en.wiktionary.org}, Verbosity \cite{von2006verbosity}, and WordNet \cite{miller1995wordnet}, to create a comprehensive network of common-sense relationships with crowd-sourced human ratings \cite{speer2013conceptnet}. Knowledge about words is represented as a semantic graph and relatedness of concepts are edges in this graph. We use these relatedness scores to construct noun--adjective associations.   

\subsubsection{Topic Modeling (LDA)}
Topic models assume that words in a document are generated from a mixture of topics, defined as probability distributions over the lexicon. We train a Latent Dirichlet Allocation (LDA) model (Blei, Ng \& Jordan, 2002) on the RCV1 news corpus \citep[804k documents]{rose2002reuters}. A noun--adjective similarity metric was obtained by computing the Euclidean distance between each word's respective distribution over topics $z$.

\subsection{Quantile normalization and correlations between metrics}
Across these seven different semantic association metrics, distributions of scores varies from Gaussian (GloVe, Word2Vec, ConceptNet5) to exponential (Bigram). To standardize scores across the set of $50$ nouns and $100$ adjectives, we used quantile normalization into a standard uniform distribution. Since metrics derived from similar model classes (e.g. vector representations) were highly correlated (Figure \ref{fig:model_macro}), we picked a subset of association metrics that derive from qualitatively different model classes with the constraint of being trained on similar corpora (e.g. news and books) whenever possible. This resulted in a choice of four measures, \textit{Bigram} (Googe Ngram), \textit{Word2Vec} (Google News), \textit{ConceptNet} (ConceptNet5), and \textit{LDA}, as candidate semantic models of human word choices.\footnote{LDA was excluded in the final two rounds of experiments. With the current training regime, its success in fitting human responses was substantially smaller than the other three semantic association measures we chose.} Unless otherwise noted, we will use `Bigram' and `Word2Vec' to refer to those metrics based on Google Ngram and Google News, respectively. 
\subsection{Optimal Experimental Design}
\label{sec:oed}
Despite focusing on a relatively small set of nouns and adjectives, the space of possible experimental configurations is still too large to allow exhaustive search. Furthermore, the model rank correlations displayed in Figure \ref{fig:model_macro} suggest that naively picking configurations could result in strongly correlated predictions. To generate experimental configurations that are highly informative with respect to discriminating between different semantic association metrics, we employed Bayesian optimal experimental design (OED) techniques \citep{cavagnaro2010adaptive}.

\begin{align}
d^* &= \argmax_{c \in D} U(c)\\
\label{eq:global_util}
U(c) &= \sum_{y} u(y, c) P(y|c)
\end{align}

Assuming that a particular response $y$ is recorded (a choice of noun pair or adjective), the utility of an experimental configuration $c$, $u(y, c)$, is proportional to the mutual information between the distributions over models $M$ before and after obtaining datum $y$. Since response $y$ has not yet been observed, we compute the expectation of $u(y, c)$ with respect to $y$ to obtain the desired (global) utility of the configuration $U(c)$. Assuming a uniform prior distribution over models $M$, the equations simplify in the following way.
Optimal designs were computed using Monte Carlo methods for sampling-based stochastic optimization \citep{muller2005simulation}. 

\begin{figure}[t]
\centering
\includegraphics[width=0.48\textwidth]{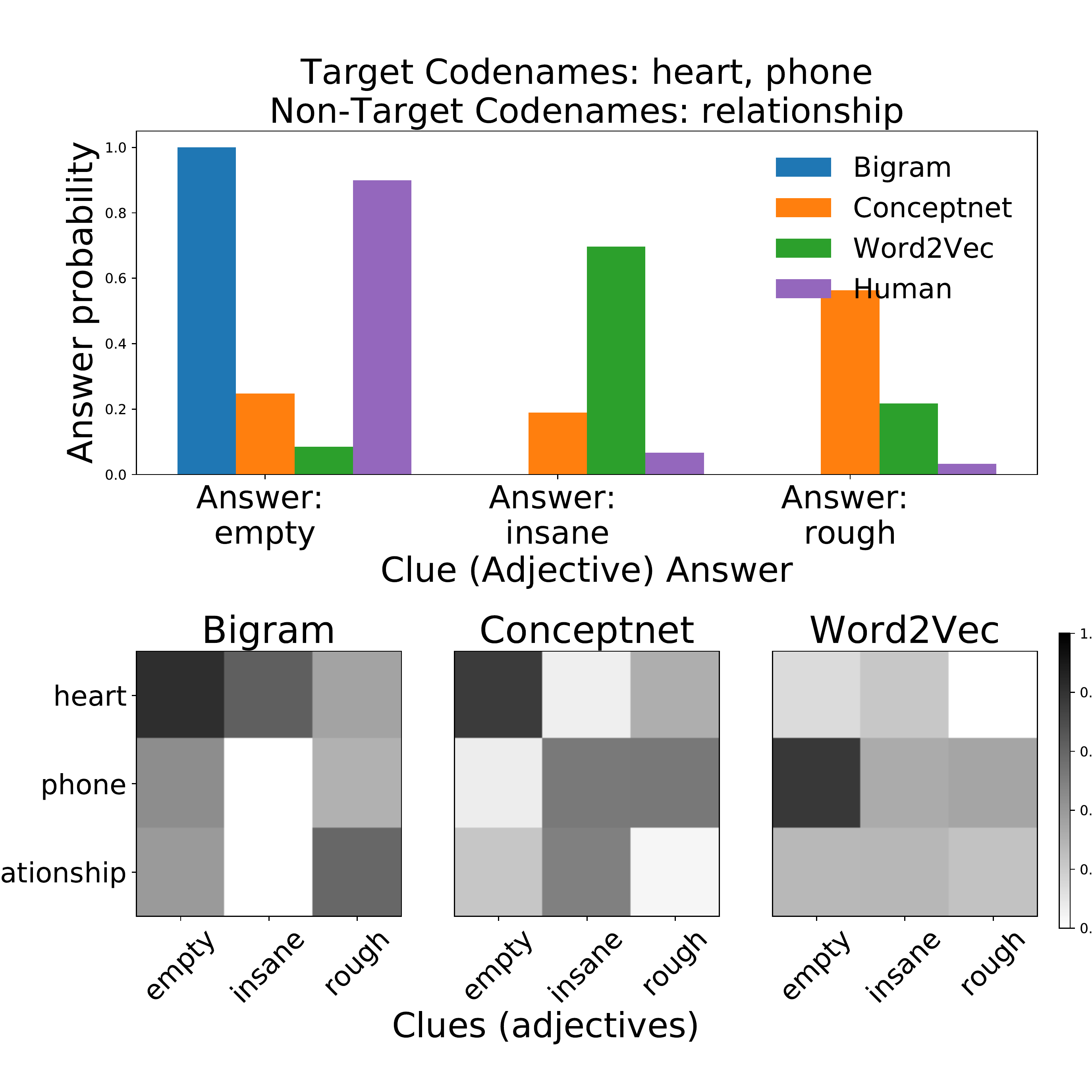}
\caption{This example speaker configuration shows how different clues are preferred by different models: `\textit{empty}' most often co-occurs with `\textit{heart}' and `\textit{phone}' and is thus favored by the Bigram model. ConceptNet assigns a high association score to `\textit{heart}' and `\textit{empty}' but the adjective `\textit{rough}' fits the noun pair better overall when using product aggregation. Similarly, since `\textit{insane}' appears most often in the context windows for both `\textit{heart}' and `\textit{phone}', it is the top prediction for Word2Vec. We also show human data for the configuration, which shows a strong preference for the adjective preferred by the Bigram model.}
\label{fig:semantic_example}
\end{figure}

Figure \ref{fig:semantic_example} illustrates a representative example configuration obtained using OED. We can see that model predictions diverge strongly, with each semantic measure predicting a different response and little distributional overlap. The accompanying matrices illustrate how the different models arrive at those predictions.

\subsection{Scoring}
To evaluate how well a particular model accounts for human responses we use two performance scores: For each configuration, we count when a model's top prediction matches the most frequent response given by participants and refer to this score as \textit{top answer}. When normalized by the total number of responses, chance performance is at $1/\#\text{answers}$. To additionally take into account information beyond the most probable answer, we computed the Spearman's \textit{rank correlation} coefficient between model predictions, sorted by probability, and subjects' choices, sorted by frequency. Chance performance is zero for this measure.

\section{Results}
\begin{table}[t!]
\small
\begin{center}
\begin{tabular}{|l|lr|lr|}
\hline 
 &\multicolumn{2}{|c|}{\bf{Top answer}} & \multicolumn{2}{|c|}{\bf{Rank correlation}} \\ 
       & Mean      & SEM & Mean      & SEM \\
\hline
\multicolumn{5}{|l|}{\bf{Listener}} \\
\hline
Bigram      &\bf 0.416 & $\pm$ 0.056 &  \bf0.384  & $\pm$ 0.037 \\
ConceptNet & 0.208 &$\pm$ 0.046    & 0.196      & $\pm$ 0.046 \\
Word2Vec    & 0.247 &$\pm$ 0.055    & 0.253      & $\pm$ 0.037 \\
LDA         & 0.052 &$\pm$ 0.050    & -0.053     & $\pm$ 0.045 \\
\hline
\multicolumn{5}{|l|}{\bf{Speaker}} \\
\hline
Bigram      & \bf 0.418 & $\pm$ 0.055      &\bf 0.516     & $\pm$ 0.033 \\
ConceptNet & 0.405& $\pm$ 0.055 &0.346      & $\pm$ 0.045 \\
Word2Vec    &  0.278 & $\pm$ 0.050    &0.279      & $\pm$ 0.043 \\
LDA         & 0.089 & $\pm$ 0.032     &0.055     & $\pm$ 0.045 \\
\hline
\end{tabular}
\end{center}
\caption{\label{tab:ex1_results} Comparison of semantic association measures in matching human responses in {Experiment 1} (No OED). Chance performance is 0.1 (listener) and 0.125 (speaker) for top answer and 0 for rank correlation.}
\end{table}

\subsection{Experiment 1: Comparing semantic metrics using heuristic designs}
In this experiment, configurations on each trial consisted of five nouns and eight adjectives. Subjects completed the task either in the speaker or in the listener condition. OED was not used for this first experiment, instead words were chosen according to heuristic criteria, detailed in the supplementary material. We did not collect confidence scores for this experiment. Using the literal speaker and listener equations from Table \ref{tab:rsa_eq} in combination with different semantic association metrics, we derived probabilistic predictions for each configuration. Predictions were scored against human responses using the two performance scores outlined above. We also explored fits of pragmatic versions of the models to the data but found that they were qualitatively similar.\footnote{Full pragmatic model fits for all experiments are reported in the supplementary material.}

Table \ref{tab:ex1_results} shows that, while all models except LDA perform above chance, the Bigram metric performs best on both the listener task and the speaker task. While the difference on the listener side is large, differences between Bigram and ConceptNet on the speaker side are substantially smaller. To gain insights into why the results for Bigram and ConceptNet were so similar we directly evaluated the models' predictions against each other, quantifying how often they make the same top prediction (\ref{fig:model_macro}C), or how rank correlated their predictions are on average (\ref{fig:model_macro}C). The bottom left matrix in Figure \ref{fig:model_macro}C shows that the measure's similarity on top answer and rank correlation on the speaker task might in part stem from their overlapping predictions. This highlights a basic design issue: The experimental designs we picked might not allow us to fully distinguish the different models by capitalizing on the differences they make in their predictions.

\subsection{Experiment 2: Comparing semantic metrics using OED}
To remedy this shortcoming and obtain better discriminability on the speaker side, we utilized optimal experiment design techniques (Section \ref{sec:oed}) to overcome the limitations associated with Experiment 1. The procedure was run for the four designated models (Bigram, Word2Vec, ConceptNet and LDA), separately for the listener and the speaker side, for $100,000$ sampling iterations. We reduced the number of nouns and adjectives to three and four, respectively, significantly decreasing search complexity. Since some high utility configurations differed only by one or two words, and some words generally occurred much more frequently than others, we eliminated configurations that differed from higher utility configurations in less than two words and by limiting the total occurrence of a word across configurations to $20$. This reduced the top $500$ configurations for each down to $119$ speaker and $137$ listener configurations. Results from prior experiments show that the difficulty of a configuration, which is not explicitly operationalized and incorporated into our search process, may significantly impact response quality. To ensure that the selected configurations generate meaningful responses from human participants, we ran a preliminary experiment on the filtered configurations and only admitted those configurations to the main experiment whose confidence rating was above mean ($58$ speaker and $67$ listener configurations).

\begin{table}[t!]
\begin{center}
\small
\begin{tabular}{|l|lr|lr|}
\hline 
 &\multicolumn{2}{|c|}{\bf{Top answer}} & \multicolumn{2}{|c|}{\bf{Rank correlation}} \\ 
       & Mean      & SEM  & Mean      & SEM \\
\hline

\multicolumn{5}{|l|}{\bf{Listener}} \\
\hline
Bigram      &\bf0.561& $\pm$ 0.092 &  \bf0.618  & $\pm$ 0.044 \\
ConceptNet &0.424 &$\pm$ 0.080    & 0.164      & $\pm$ 0.092 \\
Word2Vec    &0.545 &$\pm$ 0.091    & 0.408      & $\pm$ 0.084 \\
LDA         &0.106&$\pm$ 0.040     & -0.461     & $\pm$ 0.074 \\
\hline
\multicolumn{5}{|l|}{\bf{Speaker}} \\
\hline
Bigram      &0.130& $\pm$ 0.044      &-0.006     & $\pm$ 0.068 \\
ConceptNet & \bf 0.564& $\pm$ 0.098 &0.170      & $\pm$ 0.076 \\
Word2Vec    &  0.491& $\pm$ 0.092    &\bf 0.200      & $\pm$ 0.077 \\
LDA         &0.091 & $\pm$ 0.040     &-0.083     & $\pm$ 0.069 \\
\hline
\end{tabular}
\end{center}
\caption{\label{tab:ex2_results} Comparison of semantic association measures to human data from {Experiment 2} (separate speaker and listener OED). Chance performance is 0.33 (listener) and 0.25 (speaker) for top answer and 0 for rank correlation.}
\end{table}

Model fits were again calculated using the literal speaker and listener equations in section \ref{sec:rsa_equations}. Table \ref{tab:ex2_results} summarizes how well the four semantic association measures fit human responses. For the listener task, the Bigram association metric scores marginally higher than Word2Vec in top answer but strongly outperforms other models in rank correlation. While ConceptNet (top answer) and Word2Vec (rank correlation) win on the speaker side, surprisingly, Bigram performs considerably worse than in experiment 1. In terms of task difficulty, speakers judged the task to be more difficult than listeners ($t=8.27$, $p<0.001$).

The surprisingly low performance of the Bigram model could be due to data sparsity that was systematically exploited by OED. On average, 45\% of the Bigram values for the noun--adjective associations used in the experiment, which are used to compute model predictions, were effectively zero (i.e. zero counts are quantile normalized to $1e^{-7}$). This level of sparsity is much higher than both the total set of Bigram associations (17\%) as well as in subsequent speaker configurations (30\%). In contrast, on the listener side, the percentage of values with near zero probability is similar between this set of configurations and those in later experiments. To further explore the data sparsity hypothesis, we computed model fits using bigram associations derived from the Twitter corpus, where only 5\% of speaker configurations are sparse. This raises the fit of the Bigram model to human data to $0.37$, even though the Twitter and Google Bigram features are highly correlated.

Irrespective of how much of the bad performance of the bigram model could be explained away by data sparsity, the basic asymmetry between Bigram's performance across the two experimental conditions seems to hold. One likely confound in assessing speaker and listener resources is that we searched for high utility configurations independently, and that this difference in material is driving the difference in performance. This hypothesis was directly addressed in the next experiment. 


\subsection{Experiment 3: Comparing listeners and speakers on the same scenarios}
To further investigate potential asymmetries between the speaker and the listener condition, we modified the design optimization procedure to jointly optimize the geometric mean of all speaker and listener configurations for the same scenario. Our intention was to collect data for all possible configurations of a scenario so that we could have listeners and speakers engage with the identical words. We then applied the same filtering procedure to reduce our set to $120$ scenarios ($760$ unique configurations). Here we restrict ourselves to three adjectives, matching the number of choices on the speaker side and minimizing differences in task difficulty. Due to its weak performance in the previous experiments, we eliminated LDA from the comparison set for subsequent experiments.

\begin{table}[t!]
\begin{center}
\small
\begin{tabular}{|l|lr|lr|}
\hline 
 &\multicolumn{2}{|c|}{\bf{Top answer}} & \multicolumn{2}{|c|}{\bf{Rank Correlation}} \\ 
       & Mean      & SEM &  Mean        & SEM \\
\hline
\multicolumn{5}{|l|}{\bf{Listener}} \\
\hline
Bigram          &\bf0.586& $\pm$ 0.072& \bf 0.496     & $\pm$ 0.056 \\
ConceptNet     &0.207   & $\pm$ 0.043& -0.050        & $\pm$ 0.063 \\
Word2Vec        &0.441   & $\pm$ 0.063 & 0.242        & $\pm$ 0.064 \\
\hline
\multicolumn{5}{|l|}{\bf{Speaker}} \\
\hline
Bigram        &\bf 0.505& $\pm$ 0.047  & \bf 0.280  & $\pm$ 0.062 \\
ConceptNet   &0.290& $\pm$ 0.051  & -0.061 & $\pm$ 0.066 \\
Word2Vec      &0.383    & $\pm$ 0.059  & 0.041 & $\pm$ 0.069 \\
\hline
\end{tabular}
\end{center}
\caption{\label{tab:ex3_results} Comparison of semantic association measures to human data from Experiment 3 (joint speaker-listener OED). Chance performance is 0.33 (listener and speaker) for top answer and 0 for rank correlation.}
\end{table}
Table \ref{tab:ex3_results} summarizes how well the remaining three semantic association measures fit human responses. In contrast to Experiment 2, and in line with the results from Experiment 1, we find that Bigram associations perform best in both the listener and speaker condition. This difference is more pronounced for the Rank correlation measure, where other models perform at chance with the exception of Word2Vec in the listener task. Based on this result, it appears likely that the difference in Experiment 2 was driven by choice of scenario configurations. When adding the constraint of finding scenarios that are jointly informative in discriminating between models on the speaker side and on the listener side, Bigram robustly outperforms other semantic association measures. While reducing the number of adjectives from 4 to 3 did not result in a significant decrease in difficulty, as measured by mean confidence, the difference in difficulty between speaker and listener task ($t=9.38$, $p<0.0001$) still remains significant.


\subsection{Experiment 4: Comparing literal and pragmatic models}
Since correlation matrices from the stimuli in Experiment 3 (Figure \ref{fig:model_macro}), which was only optimized to elicit differences between the semantic association metrics, shows that the literal models' predictions are highly correlated with their pragmatic counterparts, we ran another design optimization iteration to find configurations for which literal and pragmatic models strongly disagree. We restricted ourselves to the Bigram semantic association metric because it was the highest performing model in nine out of twelve cases (across the speaker/listener sides of three experiments, on two performance scores). Again, we jointly optimized over speaker and listener configurations, using the literal version of the model and the corresponding pragmatic model with $\alpha = 1$ from applying the RSA equations in Table \ref{sec:rsa_equations}. After filtering for overlap and limiting word co-occurrence as in the previous experiments, we select the highest $60$ utility scenarios later reduced to $40$ by highest mean confidence. In the experiment, we again tested each scenario in all its six configurations.
\begin{table}[t!]
\begin{center}
\small
\begin{tabular}{|l|lr|lr|}
\hline 
 & \multicolumn{2}{|c|}{\bf{Top answer}}& \multicolumn{2}{|c|}{\bf{Rank Correlation}} \\ 
        &  Mean      & SEM & Mean       & SEM\\
\hline
\multicolumn{5}{|l|}{\textbf{Listener} (Bigram)}\\
\hline
Literal  & \bf 0.744     & $\pm$ 0.079 & \bf 0.551 & $\pm$ 0.053\\
\hline
Pra. $\alpha=0.1$  & 0.470  & $\pm$ 0.046 & 0.159 & $\pm$ 0.063\\
Pra. $\alpha = 1.0$  & 0.521  & $\pm$ 0.046 & 0.184 & $\pm$ 0.065\\
Pra. $\alpha = 5.0$  & 0.547  & $\pm$ 0.046 & 0.242 & $\pm$ 0.065\\
\hline
\multicolumn{5}{|l|}{\textbf{Speaker} (Bigram)}\\
\hline
Literal  & \bf 0.652     & $\pm$ 0.074 & \bf0.378 &$\pm$ 0.057\\
\hline
Pra. $\alpha = 0.1$ & 0.478   & $\pm$ 0.046 & 0.069 &$\pm$ 0.068\\
Pra. $\alpha = 1.0$ & 0.496   & $\pm$ 0.046 & 0.105 &$\pm$ 0.066\\
Pra. $\alpha = 5.0$&  0.496   & $\pm$ 0.046 & 0.144 &$\pm$ 0.066\\
\hline
\end{tabular}
\end{center}
\caption{\label{tab:ex4_results} Comparison of pragmatic RSA models in predicting human responses in {Experiment 4}. Chance performance is 0.33 (listener and speaker) for top answer and 0 for rank correlation.}
\end{table}

Table \ref{tab:ex4_results} summarizes the top answer and rank correlation scores for literal and pragmatic models of various degrees of pragmatic behavior ($\alpha = [0.1, 1.0, 5.0]$). We do not see strongly scalar inferential behavior of the type predicted by RSA when applied to our setting. The literal model outperforms all pragmatic models by a large margin across both performance scores and experimental conditions. As before, speakers judged the task to be more difficult than listeners ($t=6.56$, $p<0.0001$).

\begin{figure}[t]
\centering
\includegraphics[width=0.48\textwidth]{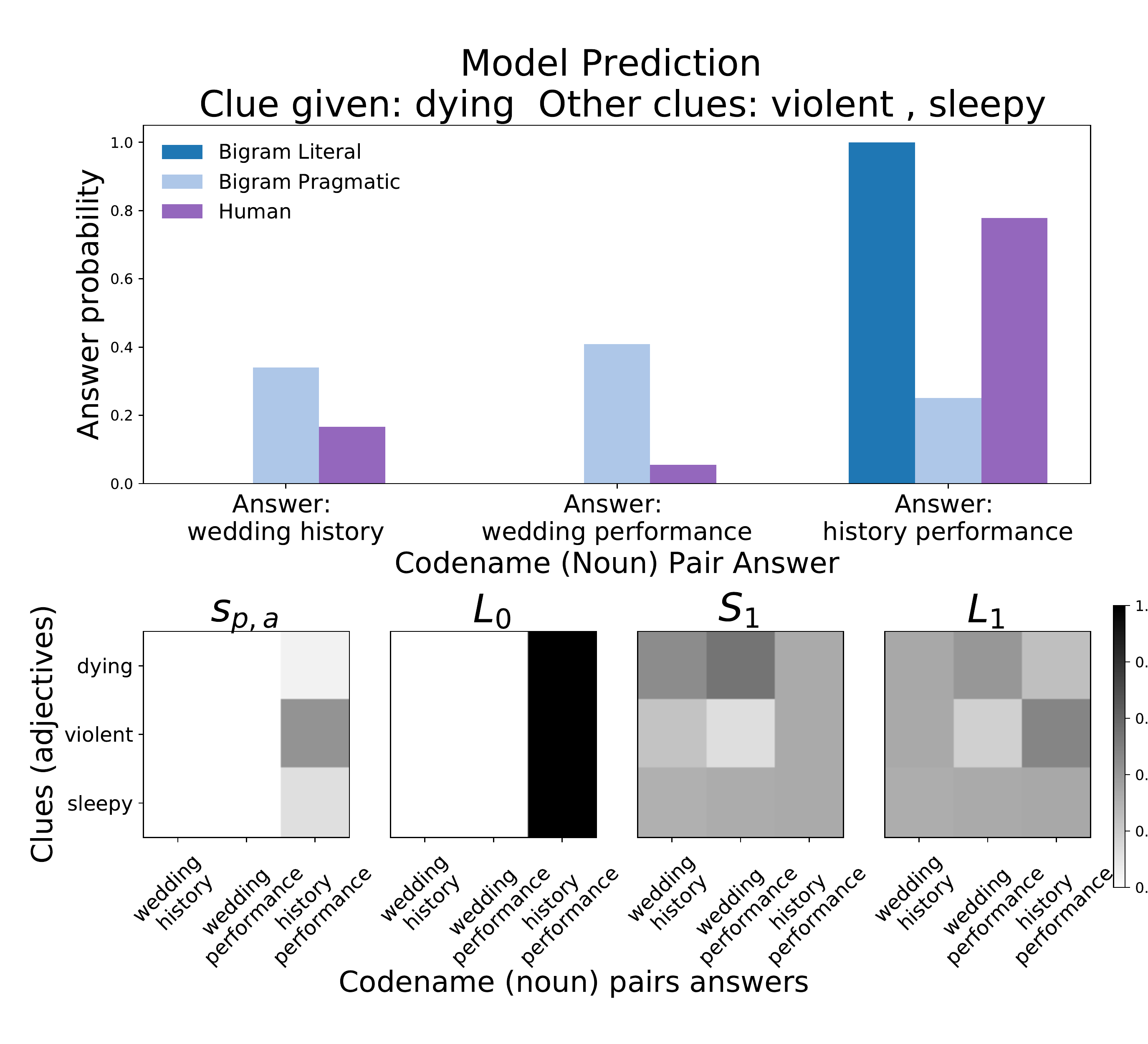}
\caption{Representative model predictions and RSA probability matrices ($\alpha =1$) from a configuration that illustrates the consequences of repeated re-normalization on model predictions.}
\label{fig:pragmatic_example}
\end{figure}

This stark difference between pragmatic and literal models is surprising. Figure \ref{fig:pragmatic_example} illustrates a common pattern that helps to better interpret this behavior. The literal model's predictions are more categorical and best reflect the probabilities from the original association values after aggregation. Through the recursive reasoning from RSA, small differences in raw probabilities, which might be non-obvious to humans, are magnified to sway top pragmatic model prediction. For example, `{history}' and `{performance}' are initially the best choice for the given adjective `{dying}' (see top row of matrices in Figure \ref{fig:pragmatic_example}), the pair is an even better choice for the adjective `{violent}'. The non-obvious advantage that `{dying}' has over `{violent}' for the pair `{wedding}' and `{performance}' is becomes dominant in the $S_1$ normalization where this pair becomes the best pair for the clue `{dying}'.\footnote{We replicated the findings with ConceptNet to find the same pattern of pragmatic reasoning over-emphasizing small differences between semantic measures which leads to poor pragmatic model performance.}

\subsection{Evaluating Human Performance} 
For experiments 1, 3, and 4, where we obtained data on matching speaker and listener scenarios, we can quantify the average one-shot success that would hold if a randomly selected speaker and listener were drawn from our experimental population and played together.  For a given scenario $G$, adjective clue $a$, speaker noun pair configuration $c$, listener choice $L$, and speaker choice $S$, the average success probability: 
\begin{align}
\sum_{a \in S} P(L=c|a,G) P(S=a|G)
\end{align}
where we use relative frequency to estimate the first term from our listener data and the second term from our speaker data.  This average success is summarized in table \ref{tab:gameplay}. This shows that even though OED (section \ref{sec:oed}) may create scenarios of a wide range of difficulties, our results as seen in Tables \ref{tab:ex3_results} and \ref{tab:ex4_results} show that our models still predict human behavior well in these difficult scenarios.

\begin{table}[t!]
\begin{center}
\small
\begin{tabular}{|l|r|r|}
\hline 
& Avg. Success & Random Success \\
\hline
Exp. 1 & 0.321  $\pm$ 0.0273 & 0.100 \\
Exp. 3 & 0.427  $\pm$ 0.0257 & 0.333 \\
Exp. 4 & 0.393  $\pm$ 0.0264 & 0.333 \\
\hline
\end{tabular}
\end{center}
\caption{\label{tab:gameplay} Summary of average success on speakers and listeners in human data.}
\end{table}

\section{Discussion}
In a series of experiments, we investigated how associative information is recruited to resolve reference in language games when truth-conditional information is not available. Experiments 1-3 compared different computational models of semantics. We found that subjects' word choices were predominantly best described using a simple bigram model, derived from Google Ngrams. Experiment 4 contrasted a literal and several pragmatic versions of the winning Bigram model and found that the literal version best fit human answers. Furthermore, despite providing speakers and listeners with the same number of alternatives to choose from, speakers consistently judged their side of the game to be harder.

While employing optimal experimental design techniques was generally helpful, especially in deriving configurations for contrasting literal and pragmatic models, the method worked to our disadvantage in experiment 2, where data sparsity in the Bigram model was exacerbated. This illustrates that, despite its strength in finding good configurations, the method might be especially prone to exploiting cases of data sparsity (where models strongly predict that a noun--adjective pair does not go together) that lead to a suboptimal choice of configurations. In future extensions of this work, taking into account uncertainty in the estimates semantic associations within the OED process could address these concerns.

With the exception of Experiment 2, our data indicate that both speaker and listener behavior are both best predicted by bigram statistics.  Experiment 4 further shows that both speaker and listener behavior are best accounted for by models without a recursive pragmatic inference component. These results are consistent with the conclusion of \citet{xu_inference} that speakers and listeners are well \emph{calibrated} to one another, bringing to bear the same lexical resource and applying it using similar principles.  

Although our experiments do not provide support for RSA as a good model of pragmatic behavior for the scenarios that most sharply distinguish level-0 and level-1 RSA models, this does not rule out the possibility that participants are not engaged in any pragmatic behavior at all. In our Experiment 4, optimal experiment design drew us to cases where pragmatic agents can transform a `least-bad' fit between a clue and target word pair to a `best' fit, through repeated renormalization of speaker and listener probability distributions.  This transformation may simply be a more arbitrary overriding of direct associative fit than humans are prepared to consider. Furthermore, there may be other types of pragmatic behavior that humans engage in for this task that we did not represent in our model space.

It is possible that, since participants in our experiments spent $20 - 30$ seconds on each question, their responses are based on first instinct while pragmatic decisions may require careful, more time-consuming reasoning. We only collected confidence ratings from participants and did not ask for their reasoning behind the answers given, thus limiting the interpretability of our findings. Another limitation is that the use of pragmatic devices in the current setup might require people to have repeated interactions so that they can align their resources more effectively. One interesting future direction of study that would make use of an interactive game design could investigate how people coordinate their reference strategies across repeated interactions.  

There are scenarios that none of the models predict correctly. This could suggest other sources of semantic information that we did not incorporate in our study. Besides competing hypotheses about the nature of the semantic knowledge deployed during the task, we suggest that the metrics could alternatively describe complementary sources of information people might draw on when playing the game. Another direction of future work could focus on combining a mixture of different semantic models in explaining human choices and should focus on factors that will likely bring out pragmatic reasoning in participants.

\section{Conclusion}
We model speaker and listener behavior through a simplified version of the game \textit{Codenames} and do not find strong evidence for the sophisticated pragmatic behavior of the type predicted by RSA-like models (Experiment 4). This suggests that there are limits on “\textit{strong}” scalar inference in one-shot associative settings. Furthermore, we find that bigram lexical statistics (Google Bigrams) were the strongest predictors of human behavior in our task, especially for listeners. This finding suggests that direct co-occurrence statistics are particularly salient in associative settings such as ours. This result may be a consequence of our restricting codenames and clues to be nouns and adjectives respectively or may hold more generally. Finally, our data suggest a potential discrepancy between the information sources relied upon by speakers and listeners: In some experiments (Experiment 2), different models performed best on the speaker and on the listener side where we would intuitively expect that successful communication requires that speakers’ and listeners’ semantic knowledge be aligned. In addition, even when controlling for the number of choices per trial, mean answer confidence in the listener condition is significantly higher, suggesting that the speaker task is intrinsically harder. Future research further exploring inference in language game settings could investigate repeated rounds of interaction, or even one-shot interaction in richer referential domains. 

\section*{Acknowledgments}

This work was supported by NSF grants BCS-1456081 and BCS-1551866 to RPL. We'd like to thank Iyad Rahwan and the Scalable Cooperation group for their valuable input and support.  

\bibliography{conll2018}
\bibliographystyle{acl_natbib_nourl}
\newpage
\newpage
\section{Supplementary Material}

\subsection{Experiment Details}
\subsubsection{Noun and Adjective Selection}
It is critical for the experiment that the participants understand the meaning of the nouns and adjectives in the game. We complied an initial list of nouns and adjectives from a well-known visual sentiment ontology~\cite{borth2013large} to make the nouns and adjectives be grounded in real-world objects. We compare these lists are against the WordNet 3.0 database~\cite{miller1995wordnet} to remove any words that have meanings in other categories than the intended one. For instance, this step would remove the adjective `\textit{sweet}' as it can also occur as a noun. Adjectives occurring often in noun bigrams such as `\textit{hot}' in `\textit{hot dog}' are removed. Lastly, we filter the list to the top nouns and adjectives as determined from our corpus of $30$B messages from the social media platform Twitter, thereby ensuring that the meaning of all the words is well-known by native speakers. We denote the set of nouns as $N$ and adjectives as $A$. The length of these twos sets are $|N| = 40$ and $|A| = 50$.

\subsection{Pragmatic Model Performance}
Table \ref{tab:ex1_prag}, \ref{tab:ex2_prag}, and \ref{tab:ex3_prag} summarize top answer accuracy and rank correlation in the first three experiments as a comparison between literal and pragmatic models. 
\begin{table*}[t!]
\centering
\begin{tabular}{|l|ll|ll|ll|ll|}
\cline{1-9}
\multicolumn{1}{|l|}{} & \multicolumn{4}{l|}{\bf{Top Answer}}                       & \multicolumn{4}{|l|}{\bf{Rank Correlation}}                         \\ \cline{2-9} 
\multicolumn{1}{|l|}{} & \multicolumn{2}{l|}{Literal} & \multicolumn{2}{l|}{Pragmatic} & \multicolumn{2}{l|}{Literal} & \multicolumn{2}{l|}{Pragmatic} \\ \cline{2-9} 
& Mean & SEM & Mean & SEM & Mean & SEM & Mean & SEM \\ \hline
\multicolumn{9}{|l|}{\bf{Listener}} \\ \hline
Bigram     & 0.416 & $\pm$ 0.056 & 0.351 & $\pm$ 0.054 & 0.386  & $\pm$ 0.037 & 0.281  & $\pm$ 0.044 \\
ConceptNet & 0.208 & $\pm$ 0.046 & 0.156 & $\pm$ 0.041 & 0.196  & $\pm$ 0.046 & 0.150  & $\pm$ 0.044 \\ 
Word2Vec   & 0.247 & $\pm$ 0.049 & 0.234 & $\pm$ 0.048 & 0.253  & $\pm$ 0.037 & 0.188  & $\pm$ 0.040 \\ 
LDA        & 0.052 & $\pm$ 0.025 & 0.091 & $\pm$ 0.033 & -0.053 & $\pm$ 0.045 & -0.064 & $\pm$  0.047        \\
\hline
\multicolumn{9}{|l|}{\bf{Speaker}} \\ \hline
Bigram     & 0.380 & $\pm$ 0.054 & 0.342 & $\pm$ 0.053 & 0.474 & $\pm$ 0.032 & 0.327  & $\pm$ 0.036 \\
ConceptNet & 0.405 & $\pm$ 0.055 & 0.291 & $\pm$ 0.051 & 0.346 & $\pm$ 0.045 & 0.270  & $\pm$ 0.042 \\
Word2Vec   & 0.278 & $\pm$ 0.050 & 0.380 & $\pm$ 0.054 & 0.279 & $\pm$ 0.043 & 0.318  & $\pm$ 0.039 \\
LDA        & 0.089 & $\pm$ 0.032 & 0.114 & $\pm$ 0.036 & 0.055 & $\pm$ 0.045 & -0.070 & $\pm$  0.043 \\
\cline{1-9}
\end{tabular}
\caption{Comparison of semantic association measures across literal and pragmatic ($\alpha = 1 $) models to human responses in experiment 1. Chance performance is $0.1$ for listener top answer and $0.125$ for speaker top answer and $0$ for rank correlation}
\label{tab:ex1_prag}
\end{table*}

\begin{table*}[t!]
\centering
\begin{tabular}{|l|ll|ll|ll|ll|}
\cline{1-9}
\multicolumn{1}{|l|}{} & \multicolumn{4}{l|}{\textbf{Top Answer}}                       & \multicolumn{4}{l|}{\textbf{Rank Correlation}}                         \\ \cline{2-9} 
\multicolumn{1}{|l|}{} & \multicolumn{2}{l|}{Literal} & \multicolumn{2}{l|}{Pragmatic} & \multicolumn{2}{l|}{Literal} & \multicolumn{2}{l|}{Pragmatic} \\ \cline{2-9} 
                     & Mean & SEM & Mean & SEM & Mean & SEM & Mean & SEM \\ \hline
                     \multicolumn{9}{|l|}{\textbf{Listener}}                                                                                                                        \\ \hline
Bigram     & 0.561 & $\pm$ 0.061 & 0.591 & $\pm$ 0.061 & -0.013 & $\pm$ 0.070  & 0.126 & $\pm$ 0.068 \\
ConceptNet & 0.424 & $\pm$ 0.061 & 0.409 & $\pm$ 0.061 & 0.170  & $\pm$ 0.076  & 0.093 & $\pm$ 0.076 \\
Word2Vec   & 0.545 & $\pm$ 0.061 & 0.545 & $\pm$ 0.061 & 0.200  & $\pm$ 0.077  & 0.231 & $\pm$ 0.074 \\ 
LDA        & 0.106 & $\pm$ 0.038 & 0.106 & $\pm$ 0.038 & -0.083 & $\pm$ 0.069  & -0.023 & $\pm$  0.072 \\ \hline
\multicolumn{9}{|l|}{\textbf{Speaker}}\\ \hline
Bigram     & 0.148 & $\pm$ 0.047 & 0.315 & $\pm$ 0.061 & 0.618  & $\pm$ 0.044  & 0.553 & $\pm$ 0.059 \\
ConceptNet & 0.481 & $\pm$ 0.066 & 0.315 & $\pm$ 0.061 & 0.164  & $\pm$ 0.092  & 0.144 & $\pm$ 0.089 \\
Word2Vec   & 0.481 & $\pm$ 0.066 & 0.463 & $\pm$ 0.065 & 0.408  & $\pm$ 0.084  & 0.309 & $\pm$ 0.080 \\
LDA        & 0.130 & $\pm$ 0.044 & 0.167 & $\pm$ 0.049 & -0.461 & $\pm$ 0.074  & -0.403 & $\pm$ 0.075     \\
\hline
\end{tabular}
\caption{Comparison of semantic association measures across literal and pragmatic ($\alpha = 1 $) models to human responses in experiment 2 (separate speaker and listener OED). Chance performance is $0.33$ for listener top answer and $0.25$ for speaker top answer and $0$ for rank correlation}
\label{tab:ex2_prag}
\end{table*}

\begin{table*}[t!]
\centering
\begin{tabular}{|l|ll|ll|ll|ll|}
\cline{1-9}
\multicolumn{1}{|l|}{} & \multicolumn{4}{l|}{\textbf{Top Answer}}                       & \multicolumn{4}{l|}{\textbf{Rank Correlation}}                         \\ \cline{2-9} 
\multicolumn{1}{|l|}{} & \multicolumn{2}{l|}{Literal} & \multicolumn{2}{l|}{Pragmatic} & \multicolumn{2}{l|}{Literal} & \multicolumn{2}{l|}{Pragmatic} \\ \cline{2-9} 
                      & Mean & SEM & Mean & SEM & Mean & SEM & Mean & SEM \\ \hline
\multicolumn{9}{|l|}{\textbf{Listener}} \\ \hline
Bigram     & 0.537 & $\pm$ 0.047 & 0.541 & $\pm$ 0.047 & 0.496  & $\pm$ 0.056 & 0.335  & $\pm$ 0.062 \\
ConceptNet & 0.243 & $\pm$ 0.041 & 0.243 & $\pm$ 0.041 & -0.050 & $\pm$ 0.063 & -0.014 & $\pm$ 0.068 \\
Word2Vec   & 0.433 & $\pm$ 0.047 & 0.351 & $\pm$ 0.045 & 0.242  & $\pm$ 0.064 & 0.127  & $\pm$ 0.066 \\ \hline
\multicolumn{9}{|l|}{\textbf{Speaker}} \\ \hline
Bigram     & 0.427 & $\pm$ 0.047 & 0.439 & $\pm$ 0.047 & 0.254  & $\pm$ 0.065 & 0.232  & $\pm$ 0.064  \\
ConceptNet & 0.313 & $\pm$ 0.044 & 0.299 & $\pm$ 0.043 & -0.061 & $\pm$ 0.066 & -0.079 & $\pm$ 0.066  \\
Word2Vec   & 0.378 & $\pm$ 0.046 & 0.411 & $\pm$ 0.047 & 0.048  & $\pm$ 0.069 & 0.090  & $\pm$ 0.072   \\
\hline
\end{tabular}
\caption{Comparison of semantic association measures across literal and pragmatic ($\alpha = 1 $) models to human responses in experiment 3 (joint speaker-listener OED). Chance performance is $0.33$ for top answer and $0$ for rank correlation}
\label{tab:ex3_prag}
\end{table*}

\end{document}


\appendix
\section{Supplementary Material}
\subsection{Experiment Details}
\subsubsection{Noun and Adjective Selection}
It is critical for the experiment that the participants understand the meaning of the nouns and adjectives in the game. We complied an initial list of nouns and adjectives from a well-known visual sentiment ontology~\cite{borth2013large} to make the nouns and adjectives be grounded in real-world objects. We compare these lists are against the WordNet 3.0 database~\cite{miller1995wordnet} to remove any words that have meanings in other categories than the intended one. For instance, this step would remove the adjective `\textit{sweet}' as it can also occur as a noun. Adjectives occurring often in noun bigrams such as `\textit{hot}' in `\textit{hot dog}' are removed. Lastly, we filter the list to the top nouns and adjectives as determined from our corpus of $30$B messages from the social media platform Twitter, thereby ensuring that the meaning of all the words is well-known by native speakers. We denote the set of nouns as $N$ and adjectives as $A$. The length of these twos sets are $|N| = 40$ and $|A| = 50$.



\subsection{Pragmatic Model Performance}
The corresponding pragmatic model results from Experiments 1-3 are summarized below (Table \ref{tab:ex1_prag}, \ref{tab:ex2_prag}, \ref{tab:ex3_prag})
\begin{table*}[t!]
\centering
\begin{tabular}{|l|ll|ll|ll|ll|}
\cline{1-9}
\multicolumn{1}{|l|}{} & \multicolumn{4}{l|}{Top Answer}                       & \multicolumn{4}{|l|}{Rank Correlation}                         \\ \cline{2-9} 
\multicolumn{1}{|l|}{} & \multicolumn{2}{l|}{Literal} & \multicolumn{2}{l|}{Pragmatic} & \multicolumn{2}{l|}{Literal} & \multicolumn{2}{l|}{Pragmatic} \\ \cline{2-9} 
                      & Accuracy       & SEM         & Accuracy         & SEM         & Accuracy       & SEM         & Accuracy        & SEM          \\ \hline
\multicolumn{9}{|l|}{Listener}                                                                                                                        \\ \hline
Bigram                & 0.416          & $\pm$ 0.056      & 0.351            & $\pm$ 0.054       & 0.386          & $\pm$ 0.037       & 0.281           & $\pm$ 0.044        \\
ConceptNet            & 0.208          & $\pm$ 0.046       & 0.156            & $\pm$ 0.041       & 0.196          & $\pm$ 0.046       & 0.150           & $\pm$ 0.044        \\
Word2Vec              & 0.247          & $\pm$ 0.049       & 0.234            & $\pm$ 0.048       & 0.253          & $\pm$ 0.037       & 0.188           & $\pm$ 0.040        \\
LDA                   & 0.052          & $\pm$ 0.025       & 0.091            & $\pm$ 0.033       & -0.053         & $\pm$0.045       & -0.064          & $\pm$0.047        \\ \hline
\multicolumn{9}{|l|}{Speaker}                                                                                                                         \\ \hline
Bigram                & 0.380          & $\pm$0.054       & 0.342            & $\pm$0.053       & 0.474          & $\pm$0.032       & 0.327           & $\pm$0.036        \\
ConceptNet            & 0.405          & $\pm$0.055       & 0.291            & $\pm$0.051       & 0.346          & $\pm$0.045       & 0.270           & $\pm$0.042        \\
Word2Vec              & 0.278          & $\pm$0.050       & 0.380            & $\pm$0.054       & 0.279          & $\pm$0.043       & 0.318           & $\pm$0.039        \\
LDA                   & 0.089          & $\pm$0.032       & 0.114            & $\pm$0.036       & 0.055          & $\pm$0.045       & -0.070          & $\pm$0.043       \\
\cline{1-9}
\end{tabular}
\caption{ \small Experiment 1 Literal and Pragmatic Model Predictions}
\label{tab:ex1_prag}
\end{table*}

\begin{table*}[t!]
\centering

\begin{tabular}{|l|ll|ll|ll|ll|}
\cline{1-9}
\multicolumn{1}{|l|}{} & \multicolumn{4}{l|}{Top Answer}                       & \multicolumn{4}{l|}{Rank Correlation}                         \\ \cline{2-9} 
\multicolumn{1}{|l|}{} & \multicolumn{2}{l|}{Literal} & \multicolumn{2}{l|}{Pragmatic} & \multicolumn{2}{l|}{Literal} & \multicolumn{2}{l|}{Pragmatic} \\ \cline{2-9} 
                      & Accuracy       & SEM         & Accuracy        & SEM          & Accuracy       & SEM         & Accuracy        & SEM          \\ \hline
\multicolumn{9}{|l|}{Listener}                                                                                                                        \\ \hline
Bigram                & 0.561          & $\pm$0.061       & 0.591           & $\pm$0.061        & -0.013         & $\pm$0.070       & 0.126           & $\pm$0.068        \\
ConceptNet            & 0.424          & $\pm$0.061       & 0.409           & $\pm$0.061        & 0.170          & $\pm$0.076       & 0.093           & $\pm$0.076        \\
Word2Vec              & 0.545          & $\pm$0.061       & 0.545           & $\pm$0.061        & 0.200          & $\pm$0.077       & 0.231           & $\pm$0.074        \\
LDA                   & 0.106          & $\pm$0.038       & 0.106           & $\pm$0.038        & -0.083         & $\pm$0.069       & -0.023          & $\pm$0.072        \\ \hline
\multicolumn{9}{|l|}{Speaker}                                                                                                                         \\ \hline
Bigram                & 0.148          & $\pm$0.047       & 0.315           & $\pm$0.061        & 0.618          & $\pm$0.044       & 0.553           & $\pm$0.059        \\
ConceptNet            & 0.481          & $\pm$0.066       & 0.315           & $\pm$0.061        & 0.164          & $\pm$0.092       & 0.144           & $\pm$0.089        \\
Word2Vec              & 0.481          & $\pm$0.066       & 0.463           & $\pm$0.065        & 0.408          & $\pm$0.084       & 0.309           & $\pm$0.080        \\
LDA                   & 0.130          & $\pm$0.044       & 0.167           & $\pm$0.049        & -0.461         & $\pm$0.074       & -0.403          & $\pm$0.075     \\
\hline
\end{tabular}
\caption{\small Experiment 2 Literal and Pragmatic Model Predictions}
\label{tab:ex2_prag}
\end{table*}

\begin{table*}[t!]
\centering
\begin{tabular}{|l|ll|ll|ll|ll|}
\cline{1-9}
\multicolumn{1}{|l|}{} & \multicolumn{4}{l|}{Top Answer}                       & \multicolumn{4}{l|}{Rank Correlation}                         \\ \cline{2-9} 
\multicolumn{1}{|l|}{} & \multicolumn{2}{l|}{Literal} & \multicolumn{2}{l|}{Pragmatic} & \multicolumn{2}{l|}{Literal} & \multicolumn{2}{l|}{Pragmatic} \\ \cline{2-9} 
                      & Accuracy       & SEM         & Accuracy        & SEM          & Accuracy       & SEM         & Accuracy        & SEM          \\ \hline
\multicolumn{9}{|l|}{Listener}                                                                                                                        \\ \hline
Bigram                & 0.537          & $\pm$0.047       & 0.541           & $\pm$0.047        & 0.496          & $\pm$0.056       & 0.335           & $\pm$0.062        \\
ConceptNet            & 0.243          & $\pm$0.041       & 0.243           & $\pm$0.041        & -0.050         & $\pm$0.063       & -0.014          & $\pm$0.068        \\
Word2Vec              & 0.433          & $\pm$0.047       & 0.351           & $\pm$0.045        & 0.242          & $\pm$0.064       & 0.127           & $\pm$0.066        \\ \hline
\multicolumn{9}{|l|}{Speaker}                                                                                                                         \\ \hline
Bigram                & 0.427          & $\pm$0.047       & 0.439           & $\pm$0.047        & 0.254          & $\pm$0.065       & 0.232           & $\pm$0.064        \\
ConceptNet            & 0.313          & $\pm$0.044       & 0.299           & $\pm$0.043        & -0.061         & $\pm$0.066       & -0.079          & $\pm$0.066        \\
Word2Vec              & 0.378          & $\pm$0.046       & 0.411           & $\pm$0.047        & 0.048          & $\pm$0.069       & 0.090           & $\pm$0.072   \\
\hline
\end{tabular}
\caption{\small Experiment 3 Literal and Pragmatic Model Predictions}
\label{tab:ex3_prag}
\end{table*}
\bibliography{conll2018}
\bibliographystyle{acl_natbib_nourl}